\documentclass[a4paper]{article}

\usepackage[english]{babel}
\usepackage[utf8x]{inputenc}
\usepackage[T1]{fontenc}

\usepackage[a4paper,top=3cm,bottom=2cm,left=3cm,right=3cm,marginparwidth=1.75cm]{geometry}

\usepackage{authblk}
\usepackage{amsmath}
\usepackage{graphicx}
\usepackage[colorinlistoftodos]{todonotes}
\usepackage[colorlinks=true, allcolors=blue]{hyperref}
\usepackage[flushleft]{threeparttable}

\definecolor{color1}{RGB}{255, 80, 80}
\definecolor{color2}{RGB}{255, 0, 255}
\definecolor{color3}{RGB}{51, 102, 255}
\definecolor{color4}{RGB}{51, 204, 204}
\definecolor{color5}{RGB}{51, 204, 51}
\definecolor{color6}{RGB}{255, 255, 0}

\usepackage[comma,authoryear]{natbib}

\title{Word Embedding based Edit Distance}
\author[1]{Yilin Niu\footnote{The work was done when the first author was an intern at ByteDance AI Lab.}}
\author[2]{Chao Qiao}
\author[2]{Hang Li}
\author[1]{Minlie Huang}
\affil[1]{\small{Department of Computer Science and Technology, Tsinghua University, Beijing, China}}
\affil[2]{\small{ByteDance AI Lab, Beijing, China}}
\affil[ ]{\small{niuyl14j@gmail.com;qiaochao@bytedance.com;lihang.lh@bytedance.com;aihuang@tsinghua.edu.cn}}

\begin{document}

\maketitle

\begin{abstract}
Text similarity calculation is a fundamental problem in natural language processing and related fields. In  recent years, deep neural networks have been developed to perform the task and high performances have been achieved. The neural networks are usually trained with labeled data in supervised learning, and creation of labeled data is usually very costly. In this short paper, we address unsupervised learning for text similarity calculation. We propose a new method called Word Embedding based Edit Distance (WED), which incorporates word embedding into edit distance. Experiments on three benchmark datasets show WED outperforms state-of-the-art unsupervised methods including edit distance, TF-IDF based cosine, word embedding based cosine, Jaccard index, etc.
\end{abstract}

\section{Introduction}

Text similarity calculation is an important problem in NLP and related fields, in which given two strings of words or symbols the similarity between them is calculated. In this paper, we consider the unsupervised setting of text similarity calculation, because in practice it is often costly to create labeled data for training of a supervised model. Without loss of generality we consider dealing with word strings (e.g., sentences) with paraphrase identification as example. In paraphrase identification, given two strings of words, we are to judge the degree of their semantic similarity.

Traditionally, unsupervised text similarity calculation is performed by using edit distance (cf., ~\citep{Navarro:2001,Jurafsky:2009}), Jaccard index (cf.,~\citep{TanSK2005}), word-embedding-based cosine similarity (cf.,~\citep{MitchellL08, milajevs2014evaluating}), etc. There are also methods which make use of additional resources such as WordNet.

In this short paper, we present a new method for text similarity calculation by combining edit distance and word embedding, referred to as word-embedding-based edit distance (WED). Although the method is not difficult to devise, there does not seem to be a study on it in the literature, as far as we know. 

WED employs both symbolic representations and vector representations for calculation of text similarity. It defines insertion, deletion, and substitution costs on the basis of both words and word embeddings. WED has the following advantages. (1) It is very easy to implement. (2) Its performance is high in the unsupervised setting. (3) It includes edit distance and Jaccard coefficient as special cases.

Experimental results on three benchmark datasets show that WED outperforms state-of-the-art methods and performs equally well with supervised methods when data size is small. The results indicate that WED is a very effective tool for unsupervised text similarity calculation. 

\section{Related Work}

There are many methods proposed for text similarity calculation. Specially, many supervised methods have been developed recently using deep neural networks~\cite{shen2014learning,hu2014convolutional,pang2016text,wan2016deep,D16-1244}. In this section, we introduce unsupervised methods for text similarity calculation, which is the focus of this paper, including symbolic, neural, and knowledge-based methods.

\subsection{Symbolic Methods}

Edit distance is a widely used measure for text similarity calculation~\citep{Navarro:2001,Jurafsky:2009}. The minimum total costs of transforming one string into the other string is computed using dynamic programming and taken as the similarity between the two strings. Edit distance takes word order into consideration in the similarity calculation. There are also variants of edit distance developed (e.g., ~\citep{ristad1998learning}).

TF-IDF based cosine is also a measure widely used for text similarity calculation~\citep{Salton:1988}. This method represents a string of words as a TF-IDF vector where each element denotes the TF-IDF score of a word. Given two strings of words, it calculates the cosine of the vectors and takes the value as the similarity between the two strings. Similarly, Jarccard index can be calculated between the two strings and utilized as similarity between them~\citep{TanSK2005}. TF-IDF based cosine and Jarccard index do not take into account word order.

\subsection{Neural Methods}

Word embedding based cosine also becomes a popular method recently. In the method, a word is represented by a vector, which is created using word embedding methods such as Word2Vec~\citep{mikolov2013efficient} and GloVe~\citep{pennington2014glove}. A string of words (i.e., sentences) is represented by a vector as well, which is obtained by summing over the word vectors of the string. Finally, cosine between the vectors of two strings is calculated and utilized~\citep{MitchellL08, milajevs2014evaluating}.

Autoencoder~\citep{hinton2006reducing} encodes a string of words into a vector (embedding) and takes it as representation of the string. The representations of strings are obtained by compression in which the strings of words are first converted into their representations by an encoder and then reconstructed by a decoder. Cosine between the embeddings of two strings given by Autoencoder can be utilized to judge whether the strings are similar.

\subsection{Knowledge based Methods}

It is also possible to utilize lexical knowledge to decide the similarity between two strings of words (e.g., sentences). WordNet is a knowledge base about word lexical relations, e.g., synonym and antonym~\citep{miller1998wordnet}. For example, one can take the minimum distance between two words in WordNet as similarity between them, and calculate the similarity between two sentences on the basis of their words~\citep{WuP94, Leacock:1998, fernando2008semantic}. 

\section{Methods}

In this section, we describe conventional edit distance (ED) and our method of word embedding based edit distance (WED).

\subsection{Edit Distance}

Let \(S_A=(w_A^1,w_A^2,\cdots ,w_A^{l_A})\) and \(S_B=(w_B^1,w_B^2,\cdots ,w_B^{l_B})\) represent two strings of words, where \(w_A^i\) and \(w_B^j\) denote the $i$-th and $j$-th words in the two strings respectively, and \(l_A\) and \(l_B\) denote the lengths of the two strings respectively. 

Edit distance defines similarity between two strings as the minimum total cost of transformation from one string into another.
There are usually three operations for the transformation, namely insertion, deletion and substitution with costs $i(\cdot)$, $d(\cdot)$ and $s(\cdot,\cdot)$ respectively. Edit distance (ED) between two strings is calculated using dynamic programming.  The cost \(c_{i,j}\) with respect to words \(w_A^i\) and \(w_B^j\) is defined as
\begin{equation}
    c_{i,j}=
    \left\{
        \begin{array}{l}
            c_{i,j-1}+i(w_B^j)\\
            c_{i-1,j}+d(w_A^i)\\
            c_{i-1,j-1}+s(w_A^i,w_B^j)
        \end{array}
    \right. 
\end{equation}
The costs of three operations are usually defined as
\begin{equation}
        \begin{array}{l}
            i(w_B^j)=1 \\
            d(w_A^i)=1 \\
            s(w_A^i,w_B^j) = \left\{
                                \begin{array}{lr}
                                     0 & w_A^i = w_B^j \\
                                     2 & w_A^i\neq w_B^j\\
                                \end{array}
                            \right. 
        \end{array}
\end{equation}

\begin{table}[htb]
\begin{center}
\begin{threeparttable}
\caption{Example of Text Similarity Calculation Using ED}
\label{tab:case-ed}
\begin{tabular}{lllllllllllllllll} \hline
How & large & is & the & largest & *       & \colorbox{color4}{city} & in & \colorbox{color5}{Alaska} & *  & *   & *   & ?&&&&\\ 
*   & *     & *  & The & *       & biggest & \colorbox{color4}{city} & in & \colorbox{color5}{Alaska} & is & how & big & ?&&&&\\ \hline
   \end{tabular}
   \begin{tablenotes}
        \small
        \item Matched content words are in same color.
    \end{tablenotes}
\end{threeparttable}
\end{center}
\end{table}

Table~\ref{tab:case-ed} gives an example of text similarity calculation using ED, where the aligned content words in dynamic programming are highlighted. We can see that ED cannot perform matching of synonyms (e.g. 'largest' vs. 'biggest') and matching of words in distance (e.g. 'how large' and 'how big'). 

\subsection{Word Embedding based Edit Distance}

We propose word embedding based edit distance (WED) to address the aforementioned problems which edit distance (ED) cannot handle.  WED calculates similarity between two strings of words based on both symbolic representations (words) and vectorial representations (embeddings) of words in the two strings.

First similarity between two words \(w_A^i\) and \(w_B^j\) is defined on the basis of both words and word embeddings 
\begin{equation}\label{eq:sim}
\begin{aligned}
    \mbox{sim}(w_A^i,w_B^j)= \left\{
                                \begin{array}{ll}
                                 1 & w_A^i = w_B^j  \\
\sigma \left(w\cdot \cos(e_A^i,e_B^j) +b \right) & w_A^i \not= w_B^j, e_A^i \mbox{\ \& } e_B^j \mbox{\ exist} \\
                                     0 & \mbox{otherwise}  \\
                                \end{array}
                            \right. 
\end{aligned}
\end{equation}
where \(e_A^i\) and \(e_B^i\) represent the embeddings of words \(w_A^i\) and \(w_B^j\) respectively, $w$ and $b$ are hyperparameters with default values of 1 and 0, $\sigma()$ denotes the sigmoid (logistic) function, and $\cos$ denotes the cosine function. The values of $\mbox{sim}(w_A^i,w_B^j)$ are in the range of $[0,1]$. Note that words may not have embeddings in practice, for example, proper nouns and numbers, and the similarity function can still be calculated in such cases. Two words are viewed as `synonyms' if the value of the $\mbox{sim}$ function between them is large enough.

The algorithm of calculating WED is the same that of ED. Both have time complexity of $O(l_A \cdot l_B)$ where $l_A$ and $l_B$ are string lengths. The only difference between WED and ED lies in the way of calculating the costs.
The cost \(c_{i,j}\) with respect to words \(w_A^i\) and \(w_B^j\) is defined as
\begin{equation}
    c_{i,j}=
    \left\{
        \begin{array}{l}
            c_{i,j-1}+ i'(w_B^j) \\
            c_{i-1,j}+ d'(w_A^i) \\
            c_{i-1,j-1}+ s'(w_A^i,w_B^j) \\
        \end{array}
    \right. 
\end{equation}
where $i'(\cdot)$, $d'(\cdot)$, and $s'(\cdot,\cdot)$ denote insertion, deletion, and substitution costs respectively. The costs of three operations are further defined as
\iffalse
\begin{equation}
        \begin{array}{l}
            i'(w_B^j)= 1 - \{\lambda \cdot \max_{w_A^i \in S_A} \mbox{sim}(w_A^i,w_B^j) + \mu \} \\
            d'(w_A^i)= 1 - \{ \lambda \cdot \max_{w_B^j \in S_B} \mbox{sim}(w_A^i,w_B^j) + \mu \} \\
             s'(w_A^i,w_B^j) = 2 - 2 \cdot \mbox{sim}(w_A^i,w_B^j) \\
        \end{array}
\end{equation}
\fi
\begin{equation}\label{eq:cost}
        \begin{array}{l}
            i'(w_B^j)= 1 - \{\lambda \cdot \max_{w_A^k \in S_A\backslash \{w_A^i\}} \mbox{sim}(w_A^k,w_B^j) + \mu \} \\
            d'(w_A^i)= 1 - \{ \lambda \cdot \max_{w_B^k \in S_B\backslash \{w_B^j\}} \mbox{sim}(w_A^i,w_B^k) + \mu \} \\
             s'(w_A^i,w_B^j) = 2 - 2 \cdot \mbox{sim}(w_A^i,w_B^j) \\
        \end{array}
\end{equation}
where $\lambda \in [0,1]$ and $\mu \in [0,1]$ are hyperparameters with default value of 1, $\mbox{sim}$ is the similarity function in (\ref{eq:sim}), $\max_{w_A^k \in S_A\backslash \{w_A^i\}} \mbox{sim}(w_A^k,w_B^j)$ denotes the largest similarity between word $w_B^j$ and a word in $S_A$ except $W_A^i$, and $\max_{w_B^k \in S_B\backslash \{w_B^j\}} \mbox{sim}(w_A^i,w_B^k)$ denotes the largest similarity between word $w_A^j$ and a word in $S_B$ except $W_B^j$.

The intuitive explanation on the cost functions is as follows. If $w_A^i$ and $w^j_B$ are synonyms, then the cost of replacing $w_A^i$ with $w^j_B$ (the substitution cost) will be very small. If there is a word $w^k_B$ in $S_B$ that is a synonym of $w_A^i$, then the cost of deleting $w_A^i$ from $S_A$ (the deletion cost) will be smaller than usual, because $w_A^i$ and $w_B^k$ may be matched from distance.  The insertion cost is defined similarly. Note that in WED one insertion plus one deletion is not equivalent to one substitution, which is the case in ED.

\begin{table}[htb]
\begin{center}
\begin{threeparttable}
\caption{Example of Text Similarity Calculation Using WED}
\label{tab:case-wed}
\begin{tabular}{llllllllllll} \hline
\colorbox{color1}{How} & \colorbox{color2}{large} & is & the & \colorbox{color3}{largest} & \colorbox{color4}{city} & in & \colorbox{color5}{Alaska} & *  & *   & *   & ?\\ 
*   & *     & *  & The & \colorbox{color3}{biggest} & \colorbox{color4}{city} & in & \colorbox{color5}{Alaska} & is & \colorbox{color1}{how} & \colorbox{color2}{big} & ?\\ \hline
   \end{tabular}
\begin{tablenotes}
    \small
    \item Matched content words are in same color.
\end{tablenotes}
\end{threeparttable}
 \end{center}
\end{table}

Table~\ref{tab:case-wed} shows text similarity calculation using WED on the same example in Table~\ref{tab:case-ed}. One can see that WED is able to handle matching of synonyms as well as matching of words/synonyms in distance.

There are only four hyperparameters in WED, namely $w$, $b$, $\lambda$, and $\mu$, which can be tuned with a development set in practice. One can easily verify that WED degenerates to ED and a variant of Jaccard under certain conditions. That means that WED considers both match in order as in ED and match in free order as in Jaccard.

\section{Experiments}

We describe our experiments in this section.

\subsection{Data}

We utilize three benchmark datasets to evaluate the performances of our method and existing methods. The evaluation metric for all methods on all datasets is accuracy. %\textbf{We use the standard split of training, development, and test sets}.

\textbf{Quora}: The dataset released from Quora\footnote{https://data.quora.com/First-Quora-Dataset-Release-Question-Pairs} contains 400k question pairs in community question answering labeled as matched or not-matched. We use the split of training, development, and test sets from previous work~\citep{WangHF17}.

\textbf{MSRP}:  The dataset for paraphrase detection released from Microsoft Research~\citep{dolan2005automatically} contains 5.8k sentence pairs, extracted from news sources on the web. Each pair has a label indicating whether the two sentences are semantically equivalent. We use the standard split of training, development, and test sets.

\textbf{CPC}: The dataset referred to as Crowdsourced Paraphrase Collection (CPC)~\citep{jiang2017understanding}, contains 2.6k manually created pairs of paraphrase sentences. In the construction, given a sentence workers are asked to write its paraphrases. We split the dataset into a development set with 626 examples and a test set with 600 examples. There is no training set for CPC.

\subsection{Experiment Procedure}

We take the following unsupervised methods as baselines: Edit Distance (ED), TF-IDF based Cosine (TF-IDF), Word Embedding based Cosine (Embedding), Jaccard index (Jaccard), Autoencoder, which are introduced in Section 2.

For reference purposes, we also apply supervised methods to the Quora and MSRP datasets. In the first method, denoted as Bag of Words(BOW), we take the summation of the word embeddings as sentence embedding. Given two sentences, we concatenate the embeddings of them, and employ a multi-layer perceptron (MLP) to decide whether they are similar, while the model is trained with labeled data. In the second method, denoted as LSTM, given two sentences, we create the embeddings of them using LSTM~\citep{lstm}, and employ an MLP to decide whether they are similar. There are two variants for both BOW and LSTM in which one layer of MLP and three layers of MLP are exploited respectively.

We tune all the parameters or hyperparameters of all methods on the development sets. Then we evaluate the results on the test sets with the optimal parameters or hyperparameters obtained from the development sets. 

\subsection{Experimental Results}

Table~\ref{tab:main_result} summarizes the results on three datasets. We can see that our method of WED outperforms all the unsupervised baseline methods in terms of accuracy. 
The best hyperparameters are in the following ranges on the three datasets $w\in (0,20]$, $b\in [-1,1]$, $\lambda \in [0,1-\mu]$, $\mu = [-2,1]$.

For MSRP, WED performs even better than supervised methods. This is because the supervised methods cannot be effectively trained with the small amount of training data. For Quora, there are about 400k examples in the training set, and thus it is possible to train supervised methods that can work better than WED.

\begin{table}[h]
    \begin{center}
    \caption{Comparison between WED and Existing Methods}
       \label{tab:main_result}
 \begin{tabular}{l|c|c|c} \hline
                    & Quora & MSRP  & CPC   \\ \hline
      ED & 0.709 & 0.699 & 0.560 \\
      TF-IDF        & 0.673 & 0.670 & 0.558 \\
      Jaccard        & 0.708 & 0.704 & 0.565 \\
      Embedding     & 0.692 & 0.704 & 0.593 \\
     Autoencoder   & 0.629 & 0.671 & 0.542 \\
      WED          & \textbf{0.718} & \textbf{0.728} & \textbf{0.672} \\ \hline
         BOW+1 Layer      & 0.720 & 0.654 & -   \\
     BOW+3 Layer    & 0.819 & 0.693 & -     \\
      LSTM+1 Layer    & 0.760 & 0.689 & -     \\
     LSTM+3 Layer     & 0.847 & 0.681 & - \\ \hline
\end{tabular}
    \end{center}
\end{table}

MSRP is a dataset widely used. The best performances of existing unsupervised methods in terms of accuracy are in the range of 0.72 and 0.74, which utilize additional knowledge such as WordNet.\footnote{https://aclweb.org/aclwiki/Paraphrase\_Identification\_(State\_of\_the\_art)} We can see that WED can perform comparably well against those methods, even it does not use any additional knowledge.

% \begin{table}[htb]
%   \centering
%   \begin{minipage}[t]{0.9\linewidth} 
%   \caption[Experiment results]{Experiment results}
%   \label{tab:results}
%     \begin{tabularx}{\linewidth}{llll}
%       \toprule[1.5pt]
%                     & Quora & MSRP  & CPC   \\\midrule[1pt]
%       BOW+1 FC\footnote{FC refers to fully connected layer.}      & 0.720 & 0.654 & -\footnote{Not evaluate on this dataset because of no enough annotated data.}     \\
%       BOW+3 FC      & 0.819 & 0.693 & -     \\
%       LSTM+1 FC     & 0.760 & 0.689 & -     \\
%       LSTM+3 FC     & \textbf{0.847} & 0.681 & -     \\\midrule[1pt]
%       Jacard        & 0.708 & 0.704 & 0.565 \\
%       WEC           & 0.692 & 0.704 & 0.593 \\
%       TF-IDF        & 0.673 & 0.670 & 0.558 \\
%       Autoencoder   & 0.629 & 0.671 & 0.542 \\
%       Edit Distance & 0.707 & 0.694 & 0.554 \\\midrule[1pt]
%       WED          & \textbf{0.715} & \textbf{0.730} & \textbf{0.673} \\
%       \bottomrule[1.5pt]
%     \end{tabularx}
%   \end{minipage}
% \end{table}

\begin{table}[h]
    \begin{center}
    \caption{Results of T-test}
       \label{tab:t-test}
 \begin{tabular}{l|c|c|c} \hline
                    & Quora & MSRP  & CPC   \\ \hline
      (WED, ED)            & 1.40 (84\%) & 1.88 (90\%) & 4.01 (99\%) \\
      (WED, TF-IDF)        & 6.92 (99\%) & 3.72 (99\%) & 4.08 (99\%) \\
      (WED, Jaccard)       & 1.56 (88\%) & 1.56 (88\%) & 3.84 (99\%) \\
      (WED, Embedding)     & 4.03 (99\%) & 1.56 (88\%) & 2.85 (99\%) \\ \hline
\end{tabular}
    \end{center}
\end{table}

To examine the significance of improvements, we conduct t-test between WED and the four unsupervised methods on the three datasets. Table~\ref{tab:t-test} shows each method pair's test statistics and probability that WED significantly outperforms the baseline.

% \begin{equation}\label{eq:t-value}
%         \begin{array}{l}
%             \triangle_i = \epsilon_i^A - \epsilon_i^B \\
%             \mu = \frac{1}{k} \sum_{i=1}^k \triangle_i \\
%             \sigma^2 = \frac{1}{k} \sum_{i=1}^k (\triangle_i - \mu) \\
%             \tau_t = |\frac{\sqrt{k}\mu}{\sigma}|
%         \end{array}
% \end{equation}

% \(\epsilon_i^A\) is the accuracy of WED on the \(i\)-th subset, while \(\epsilon_i^B\) is that of one baseline. \(\tau_t\) represents t-test result. We set \(k\) to \(11\). Table~\ref{tab:t-test} shows values of \(\tau_t\) for each method pair. We can draw a conclusion that WED significantly outperforms the four methods with probability of more than 99\%.

\subsection{Discussions}

We further conduct an experiment to see how the techniques in WED contribute to its performance, including the uses of synonym information (word embedding) and context information. In the experiment, we disable one technique and see how it affects the performance of WED. When we disable word embeddings, the value of the sim function in (\ref{eq:sim}) becomes 0 or 1, and the costs in (\ref{eq:cost}) are calculated on the basis of word exact match (referred to as `-Embedding'). When we disable comparison with words in context (set $\lambda = 0$), the costs in (\ref{eq:cost}) are only calculated on the basis of word match at the current positions (referred to as `-Context').

Table~\ref{tab:technique} summarizes the results on all three datasets. From the table, we can see that embedding and context are useful for WED, which represent main extensions from ED. Embeddings can help better capture similarity between words and context can help better leverage similar words in distance. 

\begin{table}[htb]
\begin{center}
\caption{Contribution of Techniques}
\label{tab:technique}
\begin{tabular}{lccc} \hline
                   & Quora & MSRP  & CPC  \\ \hline
    WED          & 0.718 & \textbf{0.728} & \textbf{0.672} \\
    -Embedding        & 0.715 & 0.714 & 0.608 \\
    %-Word         & 0.715 & 0.698 & 0.673 \\
     -Context     & \textbf{0.719} & 0.719 & 0.622 \\
%      -Sigmoid                  & \textbf{0.719} & 0.722 & 0.630 \\ 
\hline
   \end{tabular}
   \end{center}
\end{table}

\begin{figure}
\begin{center}
    \includegraphics[width=0.5\linewidth]{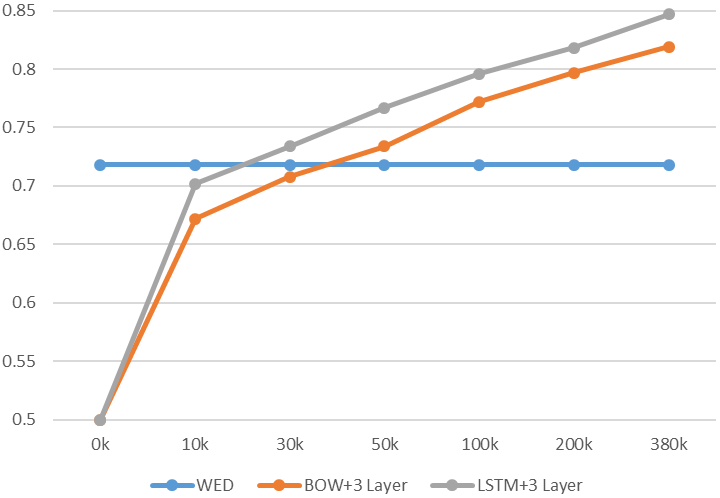}
    \caption{Performances of WED, BOW and LSTM given different sizes of data}
    \label{fig:datasize}
\end{center}
\end{figure}

We conduct an experiment to see how much training data is needed for supervised methods to achieve the same performance as WED. Figure \ref{fig:datasize} shows the results on Quora. WED is comparable with supervised methods when the data size is about 30k. Note that 30k labelled data is usually difficult to create, which means that an unsupervised method like WED is useful when there is not sufficient training data.

% \begin{figure}
% \begin{center}
%     \includegraphics[width=1.0\linewidth]{figs/error-case.PNG}
%     \caption{Error Analysis on Quora}
%     \label{fig:error-analysis}
% \end{center}
% \end{figure}
%as shown in figure~\ref{fig:error-analysis}. 

We randomly select some test examples from the Quora dataset and conduct error analysis on the data. The major type of errors (about 50\%) is due to indistinction between keywords and non-keywords in WED. If there were a mechanism to assign higher weights to the keywords, then WED would be able to achieve even higher performance. Inaccuracy in word embedding is another type of errors (18\%). Words with different meanings should have small similarities between them on the basis of their embeddings, such as "what" and "how". If not, it may lead to inaccurate matching.  Ignorance of syntactic structures of sentences leads to another sort of errors (16\%). Other types of errors (16\%) include incorrect labeling, mistreatment of irrelevant words, and no-use of external knowledge.

% \begin{table}[htb]
% \begin{center}
% \caption{Effect of Data Size}
% \label{tab:datasize}
% \begin{tabular}{lccc} \hline
%                      & BOW+3 Layer & LSTM+3 Layer & WED \\ \hline
%       380k          & 0.819 & 0.847 & 0.715 \\
%       200k          & 0.797 & 0.818 & 0.715 \\      
%       100k          & 0.772 & 0.796 & 0.715 \\     
%       50k           & 0.734 & 0.767 & 0.715 \\      
%       30k           & 0.708 & 0.734 & 0.715 \\      
%       10k           & 0.672 & 0.702 & 0.715 \\ \hline
%       \end{tabular}
%       \end{center}
% \end{table}

\section{Conclusion}

We have proposed a new method named Word Embedding based Edit Distance (WED) for unsupervised text similarity calculation. WED employs both symbolic representations and vectorial representations in calculation of similarity between two text strings. WED represents a generalization of Edit Distance and Jaccard Index. Experimental results on three benchmark datasets show that WED outperforms existing unsupervised methods and performs equally well with supervised methods when data size is small. 

There might be several possible extensions of WED in future work.  One extension would be to use external knowledge such as WordNet's synonyms to enhance the similarity calculation between words.  Another extension would be to leverage additional information to calculate importance of words (something like IDF). It would also be interesting to see whether the use of more advanced pre-training techniques proposed recently (for word embedding) would further help enhance the performance of WED.

\bibliographystyle{apalike}
\bibliography{bibtex.bib}

\end{document}